\documentclass[letterpaper, 10 pt, conference]{ieeeconf}  % Comment this line out if you need a4paper

\IEEEoverridecommandlockouts                              % This command is only needed if 
\usepackage[utf8]{inputenc}
\usepackage{amsmath}
\usepackage{mathtools}
\usepackage{amsfonts}
\usepackage{xcolor}
\usepackage[english]{babel}
\usepackage{stackengine}
\usepackage{amssymb}
\usepackage{mathrsfs}
\usepackage{algorithm, float}
\usepackage{algorithmic}
\usepackage{bbm}
\usepackage{dsfont}
\usepackage{booktabs} % for professional tables
\usepackage{caption}
\usepackage{subcaption}
\usepackage{setspace}
\usepackage{graphicx}
\usepackage[thmmarks,amsmath]{ntheorem}

\newbox{\bigpicturebox}

\newtheorem{remark}{Remark}

\theoremheaderfont{\normalfont\bfseries}
\theorembodyfont{\upshape}
\theoremstyle{nonumberbreak}

\title{\LARGE \bf
An Optimization-Based Planner with B-spline Parameterized Continuous-Time Reference Signals
}

\author{Chuyuan Tao, Sheng Cheng, Yang Zhao, Fanxin Wang and Naira Hovakimyan% <-this % stops a space
\thanks{This work is supported by NASA cooperative agreement (80NSSC22M0070), NASA ULI (80NSSC22M0070), AFOSR (FA9550-21-1-0411), NSF-AoF Robust Intelligence (2133656), and NSF SLES (2331878)}% <-this % stops a space
\thanks{$^{1}$ The authors are with the Department of Mechanical Science and Engineering, University of Illinois at Urbana-Champaign, USA.
        {\tt\small  \{chuyuan2, chengs, yz107, fanxinw2, nhovakim\} @illinois.edu}}
}

\begin{document}

\maketitle
% for submission
% \thispagestyle{empty}
% \pagestyle{empty}
% for editing
\thispagestyle{plain}
\pagestyle{plain}

%%%%%%%%%%%%%%%%%%%%%%%%%%%%%%%%%%%%%%%%%%%%%%%%%%%%%%%%%%%%%%%%%%%%%%%%%%%%%%%%

\begin{abstract}
For the cascaded planning and control modules implemented for robot navigation, the frequency gap between the planner and controller has received limited attention.
In this study, we introduce a novel B-spline parameterized optimization-based planner (BSPOP) designed to address the frequency gap challenge with limited onboard computational power in robots. The proposed planner generates continuous-time control inputs for low-level controllers running at arbitrary frequencies to track. Furthermore, when considering the convex control action sets, BSPOP uses the convex hull property to automatically constrain the continuous-time control inputs within the convex set. Consequently, compared with the discrete-time optimization-based planners, BSPOP reduces the number of decision variables and inequality constraints, which improves computational efficiency as a byproduct. Simulation results demonstrate that our approach can achieve a comparable planning performance to the high-frequency baseline optimization-based planners while demanding less computational power. Both simulation and experiment results show that the proposed method performs better in planning compared with baseline planners in the same frequency.
\end{abstract}

%%%%%%%%%%%%%%%%%%%%%%%%%%%%%%%%%%%%%%%%%%%%%%%%%%%%%%%%%%%%%%%%%%%%%%%%%%%%%%%%
\section{INTRODUCTION}
Autonomous robots have received significant interest in the past few years, with many of them being tasked to navigate in different environments autonomously. In general, the navigation task can be fulfilled with two cascaded components: high-level planners, which provide references in the form of discrete paths or control inputs; and low-level controllers, which are responsible for tracking these signals with the actuators.

For path planning problems, several algorithms have been proposed with the formulation of nonlinear optimization problems ~\cite{garcia1989model, tordesillas2021faster, singh2019end, williams2016aggressive}. However, solving these nonlinear optimization problems for low-level control actions requires significant computational resources. 
Thus, it is common~\cite{zhang2018sequential, 7400956} to simplify the planners to optimize solely in the configuration space or high-level control space, relying on low-level efficient control algorithms to track the optimized variables with actuators. 

While significant progress has been made by both planners and controllers in the field of autonomous robots, one of the issues that received limited attention is the frequency gap between low-frequency high-level planners and high-frequency low-level controllers. When solving the nonlinear optimizations formed by path planning problems, high-level planners usually provide discrete-time control input at a relatively low frequency. On the contrary, controllers can execute planned actions with considerably higher frequency. To narrow down the frequency gap, conventional optimization-based planners, which use the discrete-time control inputs as decision variables, have to introduce more decision variables into the optimization with improved frequency of the planners. However, it will significantly increase the computational load on the onboard computers. 

\begin{figure}
    \centering
    \includegraphics[width=\columnwidth]{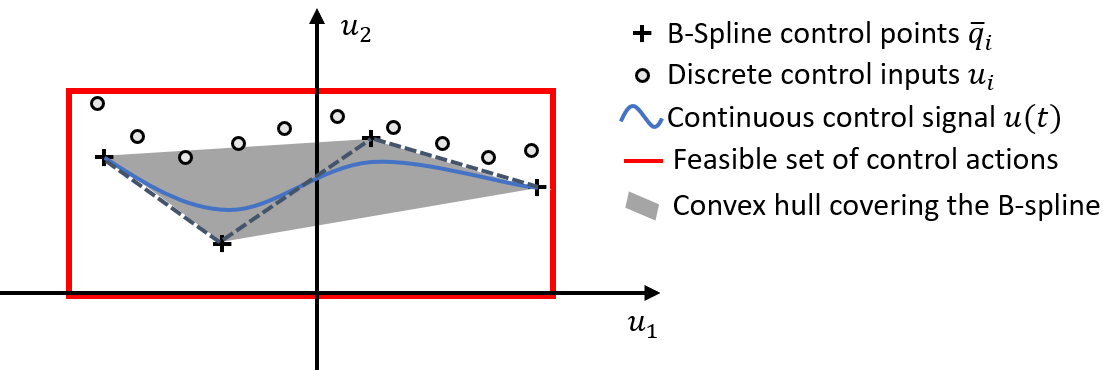}
    \caption{An illustration of discrete-time control inputs and B-spline parametrized continuous-time control signal within a convex set of control actions.} 
    \label{fig:convex-hull}
\end{figure}

In this study, we introduce a novel B-spline parameterized optimization-based planner (BSPOP), which uses B-spline to parameterize a continuous-time control input to address this challenge. The proposed planner provides continuous-time control input to address the frequency gap problem without increasing the computational burden. The B-splines used in the proposed planners are pievewise polynomial functions that can be totally determined by the control points with given time knots. In contrast to the conventional approach that makes discrete-time control actions as decision variables, in our approach, the decision variables are the control points of the B-splines. When the user requires optimal control actions at a higher frequency, the computational time of the BSPOP remains the same, since the number of optimization variables (control points) does not increase. By pre-computing the coefficients in the dynamic equations and the cost functions, BSPOP can provide a reference control signal for low-level controllers running at arbitrary frequencies to track without significantly increasing the computational burden.
Moreover, BSPOP also reduces the number of inequality constraints in optimization when considering convex control action sets. As illustrated in Figure \ref{fig:convex-hull}, by taking advantage of the convex hull property of the B-spline, the proposed planner ensures that the continuous-time control input remains within the convex control action set as long as the control points are within the same set. On the contrary, for conventional optimization-based planners, each discrete-time control input must be constrained separately in the optimization. Consequently, the proposed planner reduces the number of decision variables and inequality constraints in the optimization, which increases the computational speed and efficiency as a byproduct.

Compared with conventional discrete-time optimization-based planners, BSPOP addresses the frequency gap between planners and controllers and also decreases the number of decision variables in the optimization process. The proposed planner not only offers improved reference control inputs for arbitrarily fast low-level controllers to track but also decreases the computational demands associated with high-level baseline planners. 
We summarize our contributions as follows:
\begin{itemize}
    \item We introduce the BSPOP that generates continuous-time control signals that can be tracked by the lower-level controllers running at arbitrarily higher frequencies.
    \item We show that the BSPOP holds the number of the optimization variables invariant to the number of steps within the optimization problem's horizon. Consequently, BSPOP raises the computational speed and efficiency by reducing the number of decision variables and inequality constraints compared with a conventional planner using discrete-time control inputs as optimization variables.
    \item We validate that BSPOP can generate a better planning path compared with the same frequency baseline planner in both simulation and real-world implementation. 
\end{itemize}

The paper is structured as follows. Section \ref{sec:2} presents a review on related research. Section \ref{sec:3} outlines the formulation of the path planning problem and the baseline optimization-based planner. Section \ref{sec:4} delves into the specifics of the proposed BSPOP method. Section \ref{sec:5} presents a range of simulations and real experiments to evaluate the performance and the running speed of the proposed method compared with the baseline optimization-based planner. Finally, Section \ref{sec:6} concludes the paper and suggests potential future research directions.

\section{Related Research}\label{sec:2}
The B-spline curve is widely used to enhance the smoothness of trajectories. In previous work~\cite{zhou2019robust}, the authors constructed a B-spline curve optimization to plan fast-flight trajectories for a quadrotor, which incorporates gradient information from a Euclidean distance field and dynamic constraints. The proposed method guarantees dynamical feasibility by iteratively adjusting the non-uniform B-spline time knots. In~\cite{tordesillas2021mader}, the authors improve the B-spline basis with the MINVO basis, which provides shorter flight trajectories in multi-agent collision avoidance tasks. The authors of~\cite{wang2023speed} use the B-spline curve in the front-end search to avoid obstacles and the MINVO basis in the back-end optimization to limit the simplexes. The authors of~\cite{van2021cooperative} use the B-spline in trajectory planning for cooperative automated vehicles. Contrary to the previous work that uses the B-spline curves to parameterize the configuration space, this study uses the B-spline to parameterize continuous-time control input, allowing low-level controllers to track the control input at a user-chosen rate.

In general, most optimization-based planners are designed to provide discrete-time control inputs, which are constrained by computational power and can only provide relatively low-frequency signals. Furthermore, when using optimization-based planners, a user must balance the trade-off between performance and computation time. For a fixed planning horizon, increasing the number of discrete-time control inputs leads to improved performance~\cite{rokonuzzaman2021review, peng1991optimal}, but it also increases the number of optimization variables, which will increase the computation time.
To address this issue, continuous-time optimization-based planners have attracted the attention of both industry and researchers over the past few decades. The authors of~\cite{wang2001continuous} design orthonormal functions to parameterize the control input. In~\cite{faroni2017fast}, the authors design piecewise constant control inputs to reduce the complexity of the optimization problem. However, in the previous work, the control design does not consider constraints on the control inputs. In the proposed work, we use the convex hull property of the B-spline curve to guarantee that the control inputs are constrained within the required convex sets.

\section{Background}\label{sec:3}
\subsection{Problem Formulation of Path Planning}
We consider a nonlinear control affine system defined as:
\begin{equation}\label{eq:system}
    \dot x(t) = f(x(t)) + g(x(t)) u(t), \quad x(0)=x_0,
\end{equation}
where $x$ is the state of the system and $u$ is the control, both are defined with appropriate dimensions. We assume that the control input $u(t)$ is constrained within a known convex set that can be described by the inequality $G(u(t))\leq 0$ for all time $t$. The state of the system $x(t)$ should remain within a safe set described by $h(x(t))\geq 0$ for all time $t$. We aim to find an optimal control $u$ that can minimize a running cost function $J(x(t),u(t)) = \int_0^{T} \| x(t)- x_g \|_2^2 + u(t)^\top R u(t) dt$, where $T$ is the planning horizon, $x_g$ is the target state, and $R$ is a positive definite matrix. In summary, the path planning problem aims to find a feasible reference by solving the following optimization problem:
\begin{equation}\label{eq:continuous-opt}
    \begin{aligned}
        &\underset{u(t)}{\text{minimize}} && J(x(t),u(t)) \\
        &\text{subject to}&& \dot x(t) = f(x(t)) + g(x(t)) u(t), \\
        & && x(0) = x_c,\\
        & && G(u(t)) \leq 0, \\
        & && h(x(t)) \geq 0,\qquad t\in[0,T],
    \end{aligned}
\end{equation}
where $x_c$ is the current state of the robot.

In practical robotic applications, a common approach to solve the optimization in~\eqref{eq:continuous-opt} is to employ discretization methods to obtain approximate solutions. A conventional optimization-based planner uses the discrete-time state $x_k$ and control input $u_k$ to form the following optimization at each time step $k$:
\begin{equation}\label{eq:mpc_optimization}
    \begin{aligned}
         &\underset{u_k}{\text{minimize}} && \sum_{k=0}^{N-1} \left( \| x_k- x_g \|_2^2 + u_k^\top R u_k \right) \\
         &\text{subject to}&&  x_{k+1} = x_k + \left( f(x_k) + g(x_k)u_k \right) \Delta t, \\
         & && x_k = x_c, \\
        & && G(u_k) \leq 0, \\
        & && h(x_k) \geq 0, \qquad k\in \{0,1,\cdots,N\},
    \end{aligned}
\end{equation}
where the parameter $\Delta t$ is the time increment to discretize the time horizon, and $N$ is the number of discretized time steps in the planning horizon $T$.
The optimization problem is typically solved using numerical solvers. The first optimal control $u_0$ is then applied to the system and the process is repeated at each time step.

Decreasing the discretization interval as much as possible improves the ability of planners to closely approximate the continuous optimal control problem. However, in real robotics implementations, limited onboard computational power is also used for sensing and other algorithms, which prevents arbitrary high-frequency implementations for conventional optimization-based planners. 
Instead, these planners are usually configured at a relatively low frequency, complemented by a high-frequency, low-level controller for reference tracking. This setup can lead to performance degradation due to the frequency gap in planners and controllers. Therefore, it is ideal to have a continuous-time optimization method to overcome the gap between the low-frequency high-level planner and the high-frequency low-level controller. In the next section, we will show our approach that uses B-spline curves to parameterize a continuous-time control input, which can be tracked by arbitrarily fast low-level controllers.

\subsection{Basics of B-Spline}
The B-spline curve is a piecewise-defined polynomial curve that is smoothly blended between control points. The recursive formulation for a B-spline curve of degree $p$ with $n+1$ control points at time $t$ is given by:
\begin{equation}
    C(t) = \sum_{i=0}^{n} N_{i,p}(t) q_i,
\end{equation}
where $C(\cdot)$ is the B-spline curve, $N_{i,p}(\cdot)$ are the B-spline basis functions, and $q_i$ is a control point in the configuration space for index $i\in \{0 ,1, \cdots, n\} $. The basis functions are recursively defined as:
\begin{equation*}
    \begin{aligned}
        N_{i,0}(t) &= \begin{cases} 1 & \text{if } \tau_i \leq t < \tau_{i+1} \\ 0 & \text{otherwise} \end{cases}, \\
        N_{i,p}(t) &= \frac{t - \tau_i}{\tau_{i+p} - \tau_i} N_{i,p-1}(t) + \frac{\tau_{i+p+1} - t}{\tau_{i+p+1} - \tau_{i+1}} N_{i+1,p-1}(t).
    \end{aligned}
\end{equation*}
These formulas define the B-spline curve in terms of the contributions from each control point, weighted by the B-spline basis functions. The degree $p$ of the B-spline determines the local support of the basis functions and influences the smoothness of the resulting curve. The time knot $\boldsymbol{\tau} = \left \{ \tau_0, \tau_1, \cdots, \tau_i, \tau_{i+1}, \cdots, \tau_m \right \}$ defines the time instances at which the basis functions are evaluated. Previous studies~\cite{zhou2019robust, van2021cooperative} used the B-spline curve within the configuration space to create reference trajectories. In this study, we instead use B-spline to represent continuous-time control signals to improve the performance of planners.

\section{B-Spline Parameterized Optimization-Based Planner}\label{sec:4}
We formulate our BSPOP, which provides a continuous-time control input to address the issue of frequency gap. The continuous-time control update law can be arbitrarily fast tracked by low-level controllers. In addition, the solution benefits from the convex hull property of B-spline curves, where the resulting B-spline parameterized control signal automatically falls within the convex set of control constraints as long as the control points are within the same set.

\subsection{B-Spline Parameterized Control Input}
In the proposed method, we define the control input $u(\cdot)$ as a piecewise polynomial in time interval $[0, T]$. Denote the matrix form of the control points by $\bar Q$, which is defined as:
\begin{equation}
    \bar Q =  \begin{bmatrix}
\bar q_0 \\
\bar q_1 \\
\vdots \\
\bar q_n
\end{bmatrix},
\end{equation}
where $\bar q_i \in \mathbb{R}^{m\times 1}$ is the control-space control point for $i \in {0, 1, \cdots, n}$.
When the number of control points $n+1$ is greater than $p$, the control points determine the curves within distinct segments. In particular, the control signal $u(\cdot)$ has a total of $S=n+1-p=m-2p$ segments in the time interval, where $n+1$ is the number of control points,  $p$ is the degree of the polynomial, and $m+1$ is the length of time knots. The more control points or the higher the degree of the polynomial, the more segments  must be taken into account, leading to a more adjustable control input, but a longer computation time. For each segment $i\in \{0,\cdots, n-p\}$, the control signal $u_{i:i+1}(t)$ is expressed as follows for $t\in [\tau_i, \tau_{i+1}]$:
\begin{equation}\label{eq:control_law}
    u^\top_{i:i+1}(t)=\underbrace{ [ 1 \quad \mathcal{T}_i(t) \quad \cdots \quad (\mathcal{T}_i(t))^p]}_{\Gamma_i(t)}  M_i\underbrace{\begin{bmatrix} \bar q_{i} \\ \bar q_{i+1} \\ \vdots \\ \bar q_{i+p} \end{bmatrix}}_{\bar Q_{i+p}^i} ,
\end{equation}
where $\bar Q_{i+p}^i$ contains the control points indexed from $i$ to $i+p$, $\mathcal{T}_i(t) = \frac{t-\tau_i}{\tau_{i+1} - \tau_i}$, and $M_i$ is the basis matrix of the B-spline curve, which is decided by the time knots of the curve and we hide the notation $p$ for convenience of presentation. The shape of the B-spline curve is predetermined by the time knots, and in BSPOP, we use uniform clamped time knots (see more details in Remark 1) for the B-spline curves.
\begin{remark}
    In order to design a more adjustable control input, we employ uniform clamped time knots, which have the same value for the first $p+1$ and last $p+1$ elements, and the remaining elements in the middle increase uniformly as follows:
    \begin{equation*}
        \underbrace{\tau_0=\dots =\tau_p}_{p+1 \text{ knots}} < \underbrace{\tau_{p+1}<\dots<\tau_{n}}_{\text{uniform increasing knots}}< \underbrace{\tau_{n+1}=\dots=\tau_{m}}_{p+1 \text{ knots}},
    \end{equation*}
    where in the formulation of the BSPOP, $\tau_0= \dots = \tau_p = 0$ and $\tau_{n+1}= \dots=\tau_m=T$.
    Uniform clamped time knots allow the generation of a B-spline curve that begins at the first control point $\bar q_0$ and ends at the last control point $\bar q_n$. It provides a more adjustable control input than other time knots.
    Based on the conclusions in~\cite{qin1998general}, we can obtain the basis matrix $M$ based on the uniform clamped time knots. Note that for different segments of the control input, the basis matrix is different. The detailed derivation of the basis matrix is shown in the Appendix \ref{sec:Appendix}.
\end{remark}

\subsection{Reformulated Objective Function}
The objective function $J_{obj}$ in the BSPOP comprises two elements. The initial component $J_{goal}$ focuses on reducing the distance to the desired position $x_g$. This cost is computed numerically using the continuous-time state $x(t)Di$ and can be expressed as:
\begin{equation}
    J_{goal} = \int_0^{T} \| x(t)- x_g \|_2^2 dt.
\end{equation}
The second part of the objective function $J_{obj}$ is designed to reduce the amount of continuous-time control effort required to control the system. Given that the control input in~\eqref{eq:control_law} is divided into $S$ segments, where each segment $t$ lies within the time interval knots $[\tau_{i}, \tau_{i+1}]$, the overall cost of the control effort can be decomposed into independent costs $J_{ctrl}^{i:i+1}$, expressed as:
\begin{equation}
\begin{aligned}\label{eq:8}
        J_{ctrl}^{i:i+1}=&\int_{\tau_i}^{\tau_{i+1}} u(t)^\top R u(t) dt \\ =&\int_{\tau_i}^{\tau_{i+1}} \Gamma_i(t) M_i \bar Q_{i+p}^iR \bar Q_{i+p}^i{^\top} M_i{^\top} \Gamma_i(t){^\top} dt.
\end{aligned}
\end{equation}
To simplify the expression of the cost function $J_{ctrl}^{i:i+1}$, we simply choose the positive definite matrix $R$ to be identity. Next we separate the term $\bar Q_{i+p}^i \bar Q_{i+p}^i{^\top}$ from the time-dependent term $\Gamma_i(t)$, which permits precomputation of the time-integrated term in~\eqref{eq:8}.

Since $J_{ctrl}^{i:i+1}$ is a scalar, we start with applying to the right-hand side of~\eqref{eq:8} without changing its value as follows:
\begin{equation*}
    \begin{aligned}
        J_{ctrl}^{i:i+1} &= \text{vec}(J_{ctrl}^{i:i+1}) \\
        &= \text{vec} \left ( \int_{\tau_i}^{\tau_{i+1}} \Gamma_i(t) M_i \bar Q_{i+p}^i \bar Q_{i+p}^i{^\top} M_i^\top \Gamma_i(t){^\top} dt \right ).
    \end{aligned}
\end{equation*}
Because vectorization and integration are both linear operations, we alter their order, leading to the following result:
\begin{equation*}
    \begin{aligned}
        J_{ctrl}^{i:i+1} =  \int_{\tau_i}^{\tau_{i+1}} \text{vec} \left (\Gamma_i(t) M_i \bar Q_{i+p}^i \bar Q_{i+p}^i{^\top} M_i^\top \Gamma_i(t){^\top} \right ) dt.
    \end{aligned}
\end{equation*}
Since the control points $\bar Q_{i+p}^i$ are independent of time $t$, we use the matrix multiplication vectorization method specified in the Appendix \ref{sec:Appendix_vect} to rearrange the order of matrix multiplication and separate the control points from other terms. Consequently, we obtain the following:
\begin{equation}
\begin{aligned}
        J_{ctrl}^{i:i+1} &= \int_{\tau_i}^{\tau_{i+1}} \left(\Gamma_i(t) M_i \right ) \otimes \left ( \Gamma_i(t) M_i\right)\text{vec}(\bar Q_{i+p}^i \bar Q_{i+p}^i{^\top}) dt,  \\
         &=\underbrace{\int_{\tau_i}^{\tau_{i+1}} \left(\Gamma_i(t) M_i \right ) \otimes \left ( \Gamma_i(t) M_i \right ) dt}_{\Lambda^{i:i+1}} \text{vec}(\bar Q_{i+p}^i \bar Q_{i+p}^i{^\top}), 
\end{aligned}
\end{equation}
where $\Lambda^{i:i+1}$ is a $1 \times p^2$ row vector which can be precomputed once the time knots $\tau_i$, $\tau_{i+1}$ and the polynomial order $p$ are given. We denote the elements inside $\Lambda^{i:i+1}$ by $[\lambda_1, \dots, \lambda_{p^2}]$.
Following the definition of control points $\bar Q_{i+p}^i$ in~\eqref{eq:control_law}, we could derive the expression of matrix $\bar Q_{i+p}^i \bar Q_{i+p}^i{^\top}$:
\begin{equation*}
    \bar Q_{i+p}^i \bar Q_{i+p}^i{^\top} =  \begin{bmatrix}
\bar q_{i} \bar q_{i}^\top & \bar q_{i} \bar q_{i+1}^\top  & \cdots  & \bar q_{i} \bar q_{i+p}^\top \\
\bar q_{i+1} \bar q_{i}^\top & \bar q_{i+1} \bar q_{i+1}^\top & \cdots & \bar q_{i+1} \bar q_{i+p}^\top  \\
\vdots & \vdots & \ddots & \vdots \\
\bar q_{i+p} \bar q_{i}^\top & \bar q_{i+p} \bar q_{i+1}^\top & \cdots & \bar q_{i+p} \bar q_{i+p}^\top \\
\end{bmatrix}.
\end{equation*}
Now the control cost can be expressed in quadratic form of the control points:
\begin{equation}
   J_{ctrl}^{i:i+1} = \sum_{j_1=0}^{p} \sum_{j_2=0}^{p} \lambda_{p j_1+j_2} \bar q_{i+j_2} \bar q_{i+j_1}^\top ,
\end{equation}
where the cost function can be interpreted as the sum of the inner products of all control points $\bar q_i$ whose coefficients are determined by $\Lambda^{i:i+1}$.
Thus, the total cost of control efforts can be presented as:
\begin{equation}
    J_{ctrl} = \sum_{i=0}^{n-p+1} J_{ctrl}^{i:i+1}.
\end{equation}

To trade off the costs between reaching the target and minimizing the control effort, we use the weights $w_1$ and $w_2$ in the objective function as tuning knobs. The objective function is then defined as:
\begin{equation}
    J_{obj} = w_1 J_{goal} + w_2 J_{ctrl}.
\end{equation}

\subsection{Constraints}
In the BSPOP, we define the following constraints: dynamic constraints, collision-free constraints, and control constraints. We assume that the dynamics and environment are completely known and have been described by known functions.

\noindent \textbf{Dynamic Constraints:} Based on the dynamic system previously defined in~\eqref{eq:system} and the control signal defined in~\eqref{eq:control_law}, we reformulate the dynamic function for time $t \in [\tau_i, \tau_{i+1}]$ to be:
\begin{equation}
    \dot x(t) =  f(x(t)) + g(x(t)) \bar Q_{i+p}^i{^\top} M_i{^\top} \Gamma_i(t)^\top.
\end{equation}
To improve computational efficiency, we can modify the dynamics to make it control affine with respect to the control points $\bar Q_{i+p}^i$. Since $g(x(t))u(t)$ is a column vector, its vectorization remains unchanged. Using the matrix multiplication vectorization method specified in the Appendix \ref{sec:Appendix_vect}, we rewrite the previous dynamic equation as:
\begin{equation*}
\begin{aligned}
    \dot x(t) &=  f(x(t)) + \text{vec} \left (g(x(t)) \bar Q_{i+p}^i{^\top} M_i{^\top} \Gamma_i(t)^\top \right ), \\
     &= f(x(t)) + \underbrace{(\Gamma_i(t)  M_i) \otimes g(x(t))}_{g'(x(t))} \text{vec}(\bar Q_{i+p}^i{^\top}),
\end{aligned}
\end{equation*}
where $g'(x(t))$ is the gain matrix associated with the vectorized control points $\text{vec}( \bar Q_{i+p}^i{^\top})$.
Note that the current time $t$ determines the values of the control point $\bar Q_{i+p}^i$ and the basis matrix $M$ in the dynamic equation. 

\noindent \textbf{Collision-Free Constraints:} Since it is assumed that the obstacle in the environment is completely known and can be represented by the function $h(x(t))$, we use the inequality constraint $h(x(t))\geq 0$ to represent the robot is within the safe area.

\noindent \textbf{Control Constraints:} We aim to solve the optimization problem while control inputs are constrained within a convex region, which is defined by the convex function $G(u(t)) \leq 0$.
Using the convex hull property of the B-spline curves, as illustrated in Figure \ref{fig:convex-hull}, we can ensure that the continuous-time control input remains within the feasible convex set as long as the control points are within the convex set. Technically, if each control point $\bar q_i \in \bar Q^i_{i+p}$ satisfies $G(\bar q_i)\leq 0$, then the control input also satisfies $G(u_{i:i+1}(t))\leq 0$. In the proposed method, the control constraints are specifically related to the control points $\bar q_i$. 

\subsection{BSPOP Formulation}
In summary, given the constraints and objective functions explained above, our optimization-based planner is formulated as follows:
\begin{equation}\label{eq:bsplinempc}
\begin{aligned}
 &\underset{\bar q_i}{\text{minimize}} && J_{obj} =w_1 J_{goal} + w_2 J_{ctrl} \\
&\text{subject to}&&  \dot x(t) =f(x(t)) + g'(x(t)) \text{vec}(\bar Q_{i+p}^i{^\top}),\\
& && x(0) = x_c,\\
& && G(\bar q_i) \leq 0, \qquad i \in \{0, \cdots, n \} \\
& && h(x(t)) \geq 0, \qquad t \in [0,T].
\end{aligned}
\end{equation}

The proposed BSPOP is different from the baseline optimization-based planner in that it seeks to optimize control points $\bar q_i$, which determines continuous-time control inputs $u(t)$, instead of discrete-time control inputs $u_i$. Thus, the optimal control input of BSPOP can be tracked by low-level controllers at an arbitrary rate. In content, the conventional baseline optimization-based planner requires increased number of optimization variables to produce optimal control input $u_i$ at a higher frequency, leading to a substantial increase in computational load and solution time. In our method, we avoid the increase in computation time using the B-spline curve based control input, of which optimization variables are decoupled from the discretized time interval $\Delta t$. 
Moreover, the convex hull property of the B-spline also helps the BSPOP to reduce the number of inequality constraints in optimization. As a result, the proposed BSPOP reduces the number of decision variables and inequality constraints and improves computational speed and efficiency compared with the baseline optimization-based planners.
Specifically, the baseline planners require $\text{dim}(u) \times (T/\Delta t)$ control decision variables, where $\text{dim}(u)$ indicates the control input dimension. In contrast, the number of control decision variables needed in BSPOP is $\text{dim}(u) \times (n+1)$, where $n+1$ represents the total number of control points. When the frequency of the planners increases for a fixed prediction time horizon $T$, the discrete time interval $\Delta t$ decreases and the baseline planners introduce additional decision variables, while the BSPOP remains the same number of decision variables.
We will show the advantages of the BSPOP in both simulations and real experiments in the next section.

\section{Results}\label{sec:5}
\subsection{Simulation}
\subsubsection{Setup}
We first implement the BSPOP and baseline optimization-based planners on a unicycle model with the following dynamics:
\begin{equation}\label{eq:uni_model}
    \begin{bmatrix}
\dot {p_x}\\ 
\dot {p_y}\\ 
\dot {\theta}
\end{bmatrix} = \begin{bmatrix}
\cos \theta & 0\\ 
\sin \theta & 0\\ 
0 & 1
\end{bmatrix}\begin{bmatrix}
v \\ 
\omega 
\end{bmatrix},
\end{equation}
where $p_x, p_y$ are the Cartesian coordinates of the vehicle, $\theta$ is the vehicles orientation, $v$ is the linear velocity, $\omega$ is the angular velocity of the unicycle model. The optimization problem aims to solve for the optimal control input $v$ and $\omega$. The optimization problem for planners is configured to predict over a time horizon $T=1$ s. The robot navigation task involves an environment that requires the robots to stay within the region for $p_y\in [-2.0,1.5]$ and to avoid circle obstacles defined by:
\begin{equation*}
   h(x(t)) =\| x(t) - x_c^i \|_2^2 - r_c^i{^2} \geq 0, \quad \text{for } i\in \{ 0, 1, \dots \},
\end{equation*}
where $x_c^i$ is the center of $i$th circle obstacle and $r_c$ is the radius of the circle obstacles. The robot starts at the initial position $x_s = [-4,0]$, while the destination is specified as $x_g=[0.5, -0.5]$. The initial orientation of the robot ranges from $[-\pi, \pi]$ at intervals of $0.1$ radians. We use a third degree polynomial $p=3$ with $n+1=4$ control points to generate the B-spline curve controls. The time knots are uniformly clamped and given as $[0,0,0,0,1,1,1,1]$. The B-spline basis matrix can be computed using~\eqref{eq:matrix_deriviation}. The specific value of the matrix $M_i$ is:
\begin{equation*}
    M_i=\begin{bmatrix}
        1 & 0 &0 & 0\\
        -3 & 3 & 0 & 0 \\
        3 & -6 & 3 & 0 \\
        -1 & 3 & -3 & 1
    \end{bmatrix}.
\end{equation*}
We employ CasADi~\cite{Andersson2019} as a solver for the optimization problems. We use the Runge-Kutta 4th integration method~\cite{ascher1998computer} to numerically determine the future states. The low-level controller is designed to be a PD controller with $400$ Hz to track the optimal control inputs. The coefficients $w_1$ and $w_2$ in the cost function are set to $10$ and $1$, respectively. All simulations are carried out on a laptop that is powered by an i7-12650H CPU.

\subsubsection{Different Control Constraints}
\begin{figure}
     \centering
     \begin{subfigure}[b]{0.235\textwidth}
         \centering
         \includegraphics[width=\textwidth]{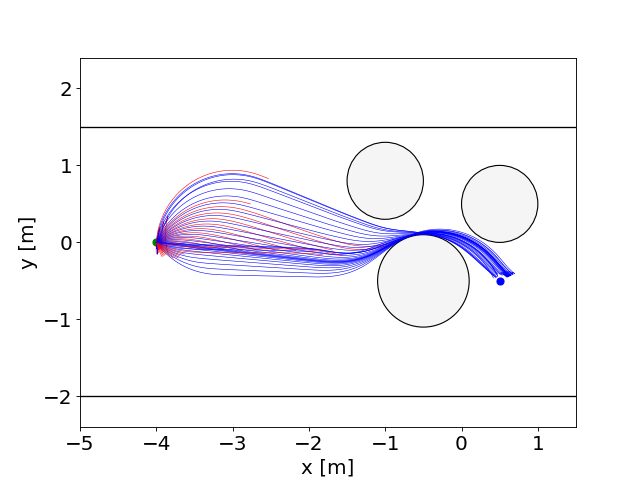}
         \caption{BSPOP trajectories.}
         \label{fig:2-1}
     \end{subfigure}
     \hfill
     \begin{subfigure}[b]{0.235\textwidth}
         \centering
         \includegraphics[width=\textwidth]{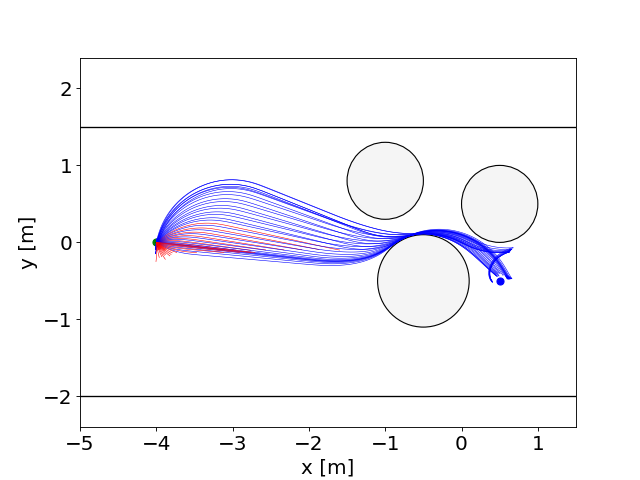}
         \caption{Baseline planner trajectories.}
         \label{fig:3-1}
     \end{subfigure}
         \caption{The performance of the BSPOP and the baseline planners under boxed control constraints.}
        \label{fig:result1}
\end{figure}

We first evaluate BSPOP and baseline optimization-based planners with boxed control constraints, where the values of $v$ and $\omega$ are constrained within a boxed region defined by $v_{\min}\leq v \leq v_{\max}$ and $\omega_{\min} \leq \omega \leq \omega_{\max}$. For this particular experiment, we set the limits for velocity as $v_{\max}=-v_{\min}=1.0$ m/s and for angular velocity as $\omega_{\max}=-\omega_{\min}=1.0$ rad/s. Both planners operate at the same frequency of $10$ Hz and utilize the same objective function. The results of the algorithms are illustrated in Figure \ref{fig:result1}. Robots that successfully reach the target are shown with blue paths, whereas the red curves indicate unfinished path where optimization problems become infeasible before the robot reaches the goal.

It is evident that both the BSPOP and the baseline optimization-based planner effectively address the navigation challenge with boxed control constraints. The BSPOP exhibits slightly superior performance, attributed to its continuous-time control parameterization and smooth control signals that prevent sharp turns near the target locations. The overall effectiveness of both planners remains comparable when operating under boxed control constraints. 

However, when considering non-trivial control constraints, the planners will have slightly different performance. For example, we consider a differential-wheeled robot which has a differential drive with two wheels of radius $r = 0.33$ m and distance $d = 0.67$ m in between. The dynamical model of the differential-wheeled robot is the same as the unicycle robot defined in~\eqref{eq:uni_model}. However, taking into account the geometry and the angular velocity limit of each wheel, we can define the following linear inequalities in the control actions $v$ and $\omega$,~\cite{tao2023difftune}:
\begin{equation*}
    \begin{aligned}
        2\omega_{\min} r \leq 2 v + \omega d \leq 2\omega_{\max} r, \\
        2\omega_{\min} r \leq 2 v - \omega d \leq 2\omega_{\max} r.
    \end{aligned}
\end{equation*}
Intuitively, these constraints restrict the vehicle from performing fast turns at high linear speed. We set the maximum angular velocity for each wheel to $\omega_{\max}=-\omega_{\min}=3.0$ rad/s. The constraints imposed define a convex diamond-shaped area for the control inputs $v$ and $\omega$.

\begin{table*}[h]
    \caption{Comparison between BSPOP and baseline optimization-based planners. The total number of control variables is shown in the last column, and the total number of decision variable is shown in the bracket.}
    \begin{center}
         \begin{tabular}{cccccc}
    \toprule[1pt]
          Planners & Comp. Time Mean (s)& Comp. Time Std (s) & CPU Usage ($\%$) &  Traj. Length (m) & Variable Num. \\ \midrule
         Baseline 10 Hz & 0.047& 0.025 & $4.6$   & 5.211 & 20 (53)\\ 
         Baseline 20 Hz & 0.077&  0.060 & $5.9$  & 5.196 & 40 (103) \\ 
         Baseline 50 Hz & 0.232& 0.240  & $6.3$   & 5.198 &100 (253) \\  
         BSPOP 10 Hz & 0.093& 0.058  & $5.0$    & 5.141 & 8 (41)\\
        \bottomrule[1pt]
    \end{tabular}   
    \end{center}
    \label{tab:table3}
\end{table*}

\begin{figure}
     \centering
          \begin{subfigure}[b]{0.235\textwidth}
         \centering
         \includegraphics[width=\textwidth]{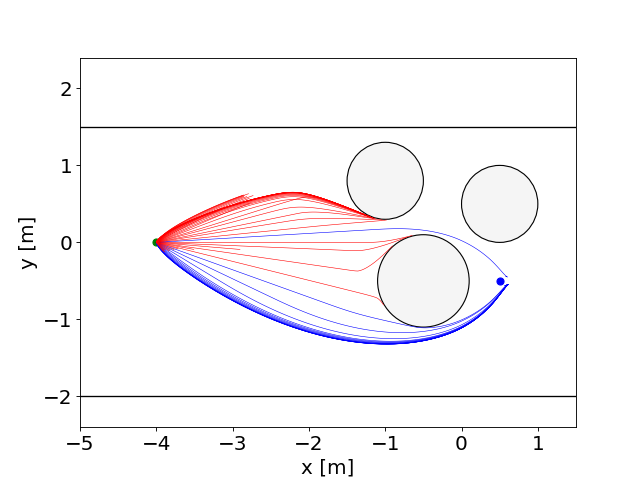}
         \caption{BSPOP trajectories.}
         \label{fig:5-1}
     \end{subfigure}
     \begin{subfigure}[b]{0.235\textwidth}
         \centering
         \includegraphics[width=\textwidth]{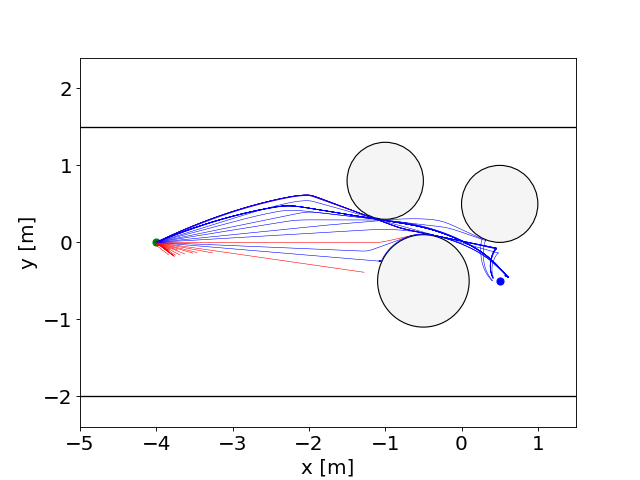}
         \caption{Baseline planner trajectories.}
         \label{fig:4-1}
     \end{subfigure}
         \caption{The performance of the BSPOP and the baseline planners with control inputs under non-trivial control constraints.}
        \label{fig:baseline_mpc2}
\end{figure}

The result of both planners is shown in Figure \ref{fig:baseline_mpc2}.
It is noted that the feasible solutions of the BSPOP approach remain continuous even when subjected to diamond-shaped constraints. On the contrary, the solutions produced by the baseline planner exhibit abrupt changes because the planner provides discrete-time control input. 
This indicates that, at equal frequencies of planners, the proposed method outperforms the conventional planners when faced with non-trivial control constraints.

% \begin{table*}[h]
%     \caption{Comparison between BSPOP and baseline optimization-based planners.}
%     \begin{center}
%          \begin{tabular}{|c|c|c|c|c|}
%     \hline
%           Planners & Comp. Time Mean (s)& Comp. Time Std (s) & CPU Usage ($\%$) &  Traj. Length (m) \\ \hline 
%          Baseline Planner 10 Hz & 0.047& 0.025 & $4.6$   & 5.211 \\ \hline
%          Baseline Planner 20 Hz & 0.077&  0.060 & $5.9$  & 5.196 \\ \hline
%          Baseline Planner 50 Hz & 0.232& 0.240  & $6.3$   & 5.198 \\  \hline
%          BSPOP 10 Hz & 0.093& 0.058  & $5.0$    & 5.141\\
%          \hline
%     \end{tabular}   
%     \end{center}
%     \label{tab:table3}
% \end{table*}

\subsubsection{Different Frequencies}
\begin{figure}
     \centering
          \begin{subfigure}[b]{0.235\textwidth}
         \centering
         \includegraphics[width=\textwidth]{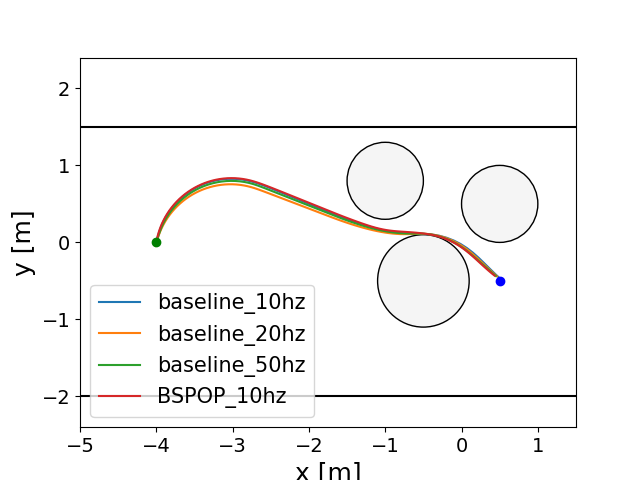}
         \caption{Trajectories of the robot subject to different planners.}
         \label{fig:-1}
     \end{subfigure}
     \hfill
     \begin{subfigure}[b]{0.235\textwidth}
         \centering
         \includegraphics[width=\textwidth]{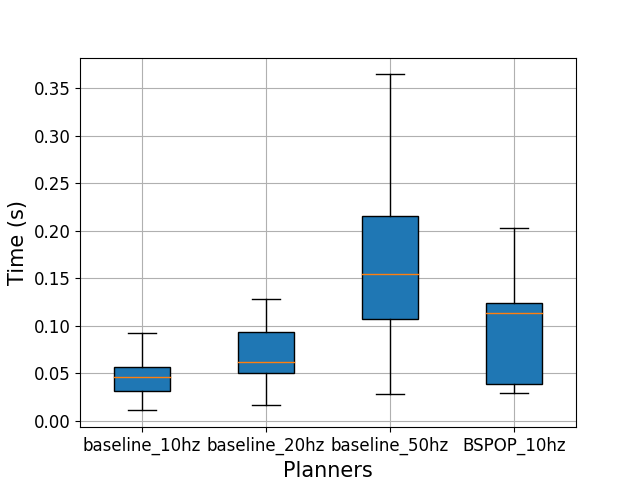}
         \caption{Boxplots of running time for different frequency planners.}
         \label{fig:-2}
     \end{subfigure}
         \caption{The performance of the baseline MPC performance when control inputs are constrained in diamond-shape area.}
        \label{fig:time_result}
\end{figure}

We compare the proposed BSPOP running in $10$ Hz and the baseline optimization-based planner running in $10, 20, 50$ Hz with the same initial state $[-4, 0, 1.4]$ and target position $[0.5, -0.5]$. We log the running time of each time step and then show their mean and standard deviation.
In addition, we track the CPU usage during the optimization process. The performance of the planners is evaluated by the total length of the trajectories. The results are shown in Table \ref{tab:table3}. From Table \ref{tab:table3}, we observe that the proposed BSPOP requires approximately $0.093$ s for computation, enabling real-time execution at a rate of $10$ Hz. The running time of the proposed method is slightly higher than the $10$~Hz baseline optimization-based planner. However, the $20$~Hz and $50$ Hz optimization-based planners take longer computation time and cannot run in real time due to the limitations of the CPU in the test.  However, these high-frequency optimization-based planners have the same performance as the proposed method from the lengths of the output trajectories. The proposed approach incurs slightly higher CPU usage compared to the $10$ Hz baseline optimization-based planner due to the computation involved in generating the B-spline curve. 
In the CasADi optimization solver used in the simulation, the decision variables comprise two parts, the state and control variables. Compared with the baseline planners, BSPOP has the same number of state variables but requires fewer control variables, since it optimizes control points instead of discrete control inputs. As shown in the Table \ref{tab:table3}, BSPOP requires fewer optimization variables, resulting in improved computational efficiency. 
The trajectory of different frequency planners is shown in Figure \ref{fig:-1}. Additionally, we illustrate the box plot showing the maximum and minimum running times of the optimization process. This visualization is presented in Figure \ref{fig:-2}.

Therefore, the BSPOP demonstrates improved performance over the baseline optimization-based planner with a same frequency, and a comparable performance compared with the high-frequency baseline planners while requiring less computational power when implemented on a computation-constrained computer onboard an autonomous vehicle.

\subsection{Experiments}
\begin{figure}
     \centering
          \begin{subfigure}[b]{0.23\textwidth}
         \centering
         \includegraphics[height=31mm]{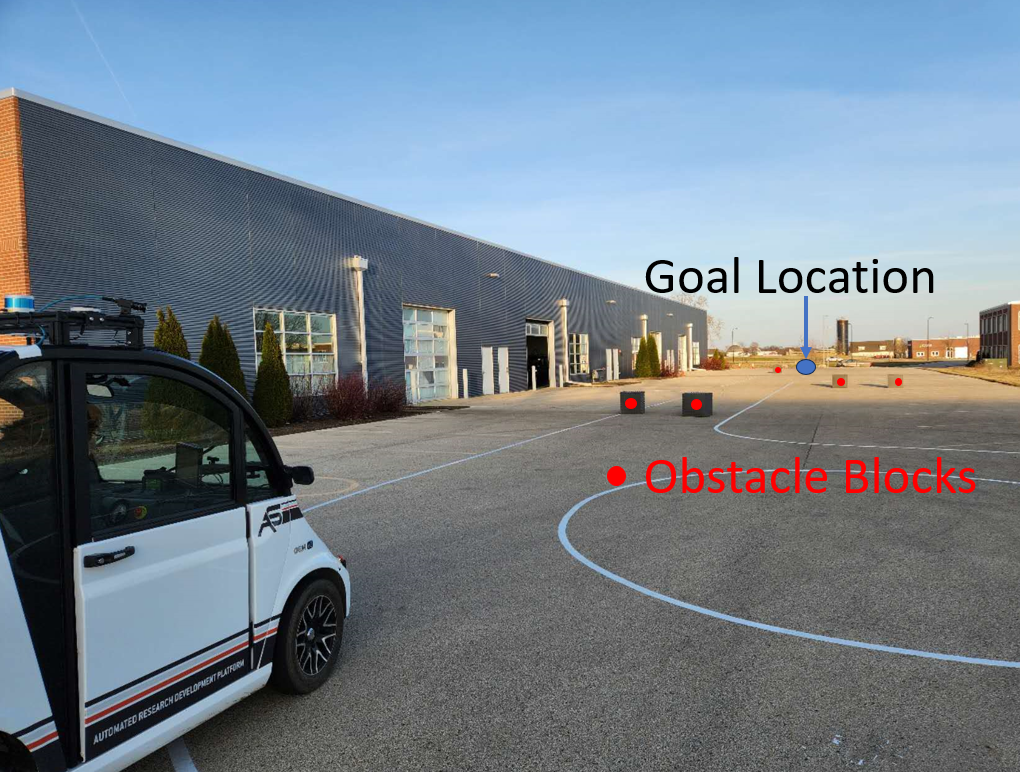}
         \caption{The environment setup of the path planning problem experiments. The vehicle is at the initial position.}
         \label{fig:realcar1}
     \end{subfigure}
     \hfill
     \begin{subfigure}[b]{0.23\textwidth}
         \centering
         \includegraphics[height=31mm]{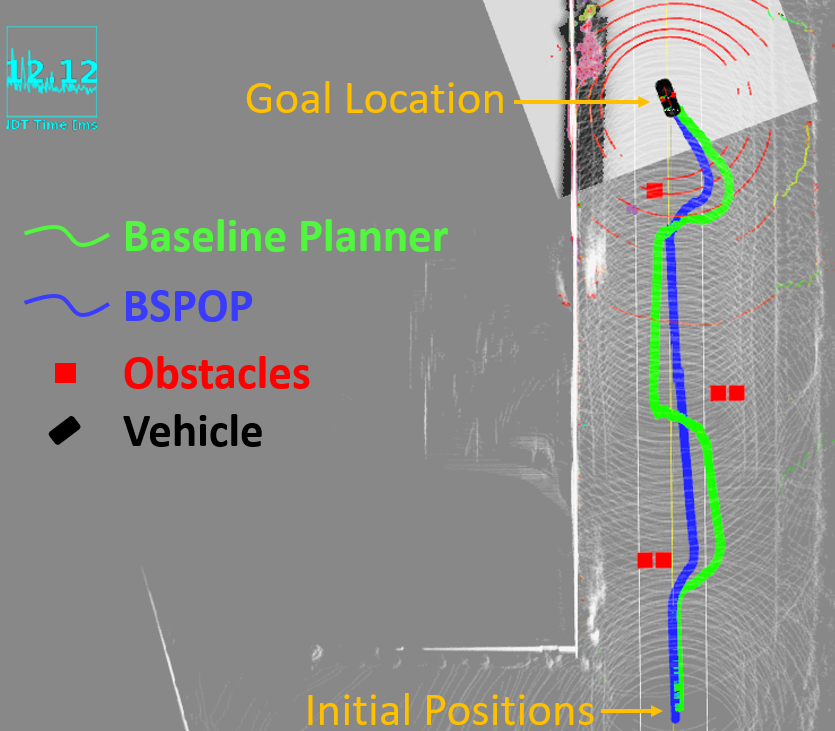}
         \caption{Trajectories of the vehicle following different planners. The vehicle is moving from bottom to top.}
         \label{fig:realcar2}
     \end{subfigure}
         \caption{Experiments using a Polaris Gem v-2 in a known environment.}
        \label{fig:real_exp}
\end{figure}

We perform the $10$ Hz BSPOP and baseline optimization-based planner on a Polaris Gem v-2 vehicle which has an Ackermann car model defined as:
\begin{equation*}
    \begin{bmatrix}
\dot {p_x}\\ 
\dot {p_y}\\ 
\dot {\theta}\\
\dot {\phi}
\end{bmatrix} = \begin{bmatrix}
\cos \theta & 0\\ 
\sin \theta & 0\\ 
\frac{\tan \phi}{L} & 0\\
0 & 1
\end{bmatrix}\begin{bmatrix}
v \\ 
\omega 
\end{bmatrix},
\end{equation*}
where $\phi$ is the steering angle and $\theta$ is the heading angle. The autonomous vehicle and the test area utilized in the experiments are shown in Figure \ref{fig:realcar1}. The vehicle uses a VLP-16 LiDAR for localization in test environments. All computation is performed onboard with an Intel Xeon E-2278G CPU. Both BSPOP and baseline optimization-based planners are executed at a frequency of $10$ Hz on board due to the sensing and mapping algorithms in Autoware~\cite{kato2018autoware}. The objective of the path planning problem is to move the car from the initial position $x_s = [0, 10]$ to the desired position $x_g = [0.2, 70]$, while the heading angle $\theta$ is initialized at $0.5$ $\pi$ and the steering angle is initialized to $0$ degrees. The obstacles in the environment are predetermined and are depicted as red blocks in Figure \ref{fig:realcar2}. The green and blue curves show the car's path when using the baseline planner and the proposed BSPOP, respectively. 
We observe that the baseline planner encounters sharp turns during driving, while the proposed planner avoids sharp turns.
Owing to the constrained computational resources onboard, higher frequency baseline planners cannot be practically executed in real-time, making it impractical to further assess them and compare them without potential safety violations.
The result of the experiments indicates that the proposed BSPOP generates a better planned path and avoids sharp turns compared with the baseline planners on the autonomous vehicle with the same frequency in the optimization setup.

\section{CONCLUSIONS}\label{sec:6}
In this study, we introduce the BSPOP, which can generate continuous-time control signals that can be tracked by lower-level controllers running at arbitrarily higher frequencies. 
We reformulate the objective function and constraints in optimization, which allows the BSPOP to hold the number of optimization variables invariant to the number of steps within the optimization problem's horizon. Besides, we show the details of the precomputable coefficients in the B-spline parameterized control input. We also analyze that BSPOP can improve computational speed and efficiency by reducing the number of decision variables and inequality constraints compared with a conventional planner using discrete-time control inputs as optimization variables.
We validate that the proposed BSPOP can provide comparable planning performance with higher-frequency planners while not significantly increasing the computational load in simulations. We also show that the BSPOP has a better planned path with the same frequency planner in both simulations and real-world experiments. The current framework assumes convex control constraints. However, the constraints in real robotic systems are often more complex. To address this, we plan to employ learning-based approaches to learn a feasible set of control actions, which will be reformulated into the optimization of the proposed planner.

\section{Acknowledgement}
The authors would like to thank the Center for Autonomy Robotics Laboratories at the University of Illinois Urbana-Champaign for their pivotal support and resources in conducting this experiment.

%%%%%%%%%%%%%%%%%%%%%%%%%%%%%%%%%%%%%%%%%%%%%%%%%%%%%%%%%%%%%%%%%%%%%%%%%%%%%%%%

%%%%%%%%%%%%%%%%%%%%%%%%%%%%%%%%%%%%%%%%%%%%%%%%%%%%%%%%%%%%%%%%%%%%%%%%%%%%%%%%

%%%%%%%%%%%%%%%%%%%%%%%%%%%%%%%%%%%%%%%%%%%%%%%%%%%%%%%%%%%%%%%%%%%%%%%%%%%%%%%%
\section*{APPENDIX}
\subsection{B-Spline Basis Matrix}\label{sec:Appendix}
The basis matrix $M_i^{p+1}$ of the B-spline basis functions of degree $p$ can be obtained by a recursive equation as follows:
\begin{equation}\label{eq:matrix_deriviation}
    M^{p+1}_i=\begin{bmatrix}
        M^{p}_i\\
        0
    \end{bmatrix} D_0 + \begin{bmatrix}
        0\\
        M^{p}_i 
    \end{bmatrix} D_1,
\end{equation}
where $D_0$ is defined as:
\begin{equation*}
 \begin{bmatrix}
1-d_{0,i-p+1} & d_{0,i-p+1} &  &  & 0 \\
 & 1-d_{0,i-p+2} & d_{0,i-p+2} &  &  \\
 &  & \ddots & \ddots &  \\
0 &  &  & 1-d_{0,i} & d_{0,i} \\
\end{bmatrix},
\end{equation*}
and $D_1$ is defined as:
\begin{equation*}
     \begin{bmatrix}
-d_{1,i-p+1} & d_{1,i-p+1} &  &  & 0 \\
 & -d_{1,i-p+2} & d_{1,i-p+2 } &  &  \\
 &  & \ddots & \ddots &  \\
0 &  &  & -d_{1,i} & d_{1,i} \\
\end{bmatrix},
\end{equation*}
where notions $d_{0,j}$ and $d_{1,j}$ are defined as follow:
\begin{equation*}
        d_{0,j} = \frac{\tau_i - \tau_j}{\tau_{j+p-1} - \tau_j}, d_{1,j} = \frac{\tau_{i+1} - \tau_j}{\tau_{j+p-1} - \tau_j},
\end{equation*}
with the convention $0/0=0$, and $M_i^0 = 1$. 

\subsection{Vectorization Formula}\label{sec:Appendix_vect}
We use the following vectorization formula~\cite{macedo2013typing}:
\begin{equation}\label{eq:vectorization}
    \text{vec}(ABC) = (C^\top \otimes A)\text{vec}(B),
\end{equation}
where the matrices $A$, $B$, and $C$ are in appropriate dimensions, $\text{vec}(\cdot)$ represents the vectorization of the matrix, and $\otimes$ is the Kronecker product.

\bibliographystyle{ieeetr}
\bibliography{Sources}

\end{document}